\documentclass[sw]{iosart2x}

\usepackage{dcolumn}
\usepackage{algorithm}
\usepackage{algpseudocode}
\usepackage[small]{caption}
\usepackage[fleqn]{amsmath}
\usepackage{graphicx}
\usepackage{slashbox}
\usepackage{subfigure}
\usepackage{multirow}
\usepackage{hyperref}
\usepackage{ulem}

\algnewcommand\algorithmicforeach{\textbf{for each}}
\algdef{S}[FOR]{ForEach}[1]{\algorithmicforeach\ #1\ \algorithmicdo}

\newcolumntype{d}[1]{D{.}{.}{#1}}

\firstpage{1}
\lastpage{5}
\volume{1}
\pubyear{0000}

\begin{document}
\begin{frontmatter} 

%

\title{Semantic Referee: A Neural-Symbolic Framework for Enhancing Geospatial Semantic Segmentation}

\runningtitle{Semantic Referee: A Neural-Symbolic Framework for Enhancing Geospatial Semantic Segmentation}

\author[A]{
\inits{F.}
\fnms{Marjan}
\snm{Alirezaie}
\ead[label=e1]{firstname.lastname@oru.se}%
\thanks{Corresponding author. \printead{e1}.}%
},
\author[A]{
\inits{S.}
\fnms{Martin}
\snm{L\"{a}ngkvist}
},
\author[B]{
\inits{T.}
\fnms{Michael}
\snm{Sioutis}
}
and
\author[A]{
\inits{FO.}
\fnms{Amy}
\snm{Loutfi}
}
\runningauthor{F. Author et al.}
\address[A]{Center for Applied Autonomous Sensor Systems, \"Orebro University, \"Orebro, Sweden}
\address[B]{Department of Computer Science, Aalto University, Espoo, Finland}

\begin{review}{editor}
\reviewer{\fnms{First} \snm{Editor}\address{\orgname{University or Company name}, \cny{Country}}}
\reviewer{\fnms{Second} \snm{Editor}\address{\orgname{University or Company name}, \cny{Country}}}
\end{review}
\begin{review}{solicited}
\reviewer{\fnms{First Solicited} \snm{Reviewer}\address{\orgname{University or Company name}, \cny{Country}}}
\reviewer{\fnms{Second Solicited} \snm{Reviewer}\address{\orgname{University or Company name}, \cny{Country}}}
\end{review}
\begin{review}{open}
\reviewer{\fnms{First Open} \snm{Reviewer}\address{\orgname{University or Company name}, \cny{Country}}}
\reviewer{\fnms{Second Open} \snm{Reviewer}\address{\orgname{University or Company name}, \cny{Country}}}
\end{review}

\begin{abstract}

Understanding why machine learning algorithms may fail is usually the task of the human expert that uses domain knowledge and contextual information to discover systematic shortcomings in either the data or the algorithm. In this paper, we propose a \textit{semantic referee}, which is able to extract qualitative features of the errors emerging from deep machine learning frameworks and suggest corrections. The semantic referee relies on ontological reasoning about spatial knowledge in order to characterize errors in terms of their spatial relations with the environment. Using semantics, the reasoner   interacts with the learning algorithm as a supervisor. In this paper, the proposed method of the interaction between a neural network classifier and a semantic referee shows how to improve the performance of semantic segmentation for satellite imagery data.

\end{abstract}

\end{frontmatter}
 
\section{Introduction}

Machine learning algorithms and Semantic Web technologies have both been widely used in geographic information systems~\cite{DBLP:journals/corr/abs-1710-03959,DBLP:conf/rweb/JanowiczSA13}. The former are typically applied on geo-data to perform computer vision tasks such as semantic segmentation, land use/cover, and object detection/recognition, whereas the latter are used for a number of applications such as navigation, knowledge acquisition and map query~\cite{surveyonto}. Recent developments in machine learning, and in particular deep learning methods, have shown large improvements for several tasks in remote sensing. However, seldom do these approaches take into account the advantages of the semantics that are associated with geo-data. Instead, the training process of neural-based algorithms typically rely on reducing an error defined by a cost function and adapt the model parameters to minimize this error, and there is no consideration if the found solution makes semantically sense. 

In the context of semantic segmentation for geospatial data, a classifier that uses only the RGB (Red, Green, Blue) channels as input is error-prone to the visual similarity between certain classes. For example, the RGB data for \textit{water} is very similar to \textit{roads} that are covered by a \textit{shadow}, and \textit{buildings} with gray roofs that are similar to \textit{roads}. One possible solution to this problem is to include additional sources of information as part of the input data to the classifier, such as Synthetic-aperture radar (SAR), Light detection and ranging (LIDAR), Digital Elevation Model (DSM), hyperspectral bands, near-infrared (NIR) bands, and synthetic spectral bands~\cite{ma2017review,cheng2017remote}. However, such additional data is not always accessible, e.g. satellite images from Google Maps or several publicly available data sets that only contain the RGB channels. One possible solution to increase the performance of the classifier is to change the architecture of the network to increase the capacity, e.g. by using Deep Convolutional Neural Networks (DCNNs)~\cite{ball2017comprehensive,zhang2017learning,guirado2017deep}. 

In this paper, instead of relying on additional sources of information or taking the ad-hoc approach of experimenting with the architecture of the classifier, we propose a method that focuses on conceptualizing the errors in terms of their spatial relations and surrounding neighborhood. Our method applies a reasoner upon an ontological representation of the context in order to retrieve the spatial and geometrical characteristics of the data. We refer to this process as a \textit{semantic referee}, since we use knowledge representation and reasoning methods to arbitrate on the errors arising from the misclassifications. 

In particular, our representation makes use of \mbox{RCC-8} spatial relations, as well as extensions thereof, where RCC-8 stands for the language that is formed by the $8$ base relations of the Region Connection Calculus~\cite{Cohn97qualitativespatial}, viz., \textit{disconnected}, \textit{externally connected}, \textit{overlaps}, \textit{equal}, \textit{tangential proper part}, \textit{non-tangential proper part}, \textit{tangential proper part inverse}, and \textit{non-tangential proper part inverse}. Notably, RCC-8 has been adopted by the GeoSPARQL\footnote{\url{http://www.opengeospatial.org/standards/geosparql}} standard, and has found its way into various Semantic Web tools and applications over the past few years~\cite{DBLP:conf/rweb/KoubarakisKKNS12}. A cross-disciplinary application of RCC-8 reasoning involves segmentation error correction for images of hematoxylin and eosin (H\&E)-stained human carcinoma cell line cultures~\cite{DBLP:journals/jimaging/RandellGFML17}. In our work, inspired by the integration of qualitative spatial reasoning methods into imaging procedures as described  by Randell et al., 2017~\cite{DBLP:journals/jimaging/RandellGFML17}, we aim to employ spatially-enhanced ontological reasoning techniques in order to assist deep learning methods for image classification via interaction and guidance.

In general, one of the key challenges in Artificial Intelligence is about reconciliation of data-driven learning methods with symbolic reasoning~\cite{Garcez2014}. The integration approaches between low and high level data have been addressed under different names depending on the employed representational models, and include abduction-induction in learning~\cite{mooney:bkchapter00}, structural alignment~\cite{conf/ic3k/AlirezaieL12}, and neural-symbolic methods~\cite{Besold2017,Bader2005}. Due to the increasing interest in deep learning methods, design and development of neural-symbolic systems has recently become the focus of different communities in Artificial Intelligence, as they are assumed to provide better insights into the learning process~\cite{Doran2017}.

\subsection{Contribution}

In this work, we develop an ontology-based reasoning approach, a preliminary version of which can be found in~\cite{lrijcai2018}, to assist a neural network classifier for a semantic segmentation task. This assistance can be used in particular to represent typical errors and extract their features that eventually assist in correcting misclassification. Using a specific case on large scale satellite data, we show how Semantic Web resources interact with deep learning models. This interaction improves the classification performance on a city wide scale, as well as a publicly available data set. 

Our contribution differentiates from the neural-symbolic systems explained in Section~\ref{sec:relatedwork} in three regards. Firstly, our method plays the role of a semantic referee for the imagery data classifier in order to conceptualize its errors, which, to the best of our knowledge, is the first attempt in the domain of image segmentation to tackle the problem by explaining its features. Secondly, our model focuses on the misclassifications and uses ontological knowledge together with a geometrical processing to explain them. This combination, to the best of our knowledge, is the first time to be employed for the aforementioned purpose. Finally, our system closes the communication loop between the classifier and the semantic referee.

\subsection{Structure of paper}

The rest of the paper is structured as follows. Section~\ref{sec:relatedwork} describes the related work. The method is presented in Section~\ref{sec:method}, which gives the overview of the approach (Section~\ref{sec:approach}), the satellite image data used in this work (Section~\ref{sec:data}), the neural network-based semantic segmentation algorithm (Section~\ref{sec:dataclassification}), OntoCity as the ontological knowledge model (Section~\ref{sec:ontocity}), and the semantic augmentation process and how it is used to guide the classifier (Sections~\ref{sec:expmis} and~\ref{sec:improve}). The experimental evaluation is presented in Section~\ref{sec:result}, which is followed by a discussion and possible directions for future work in Section~\ref{sec:discussion}.

\section{Related Work}
\label{sec:relatedwork} 

As discussed in the work done by Xie et al., 2017, in neural-symbolic systems where the learning is based on a connectionist learning system, one way of interpreting the learning process is to explain the classification outputs using the concepts related to the classifier's decision~\cite{Xie2017}. However, there is a limited body of work where symbolic techniques are used to explain the conclusions. The work presented by Hendricks et al., 2016 introduces a learning system based on a Long-term Convolutional Network (LTCN)~\cite{DBLP:journals/pami/DonahueHRVGSD17} that provides explanations over the decisions of the classifier~\cite{Hendricks2016}. An explanation is in the form of a justification text. In order to generate the text, the authors have proposed a loss function upon sampled concepts that, by enforcing global sentence constraints, helps the system to construct sentences based on discriminating features of the objects found in the scene. However, no specific symbolic representation was provided, and the features related to the objects are taken from the sentences that are already available for each image in the dataset (CUB dataset~\cite{WahCUB_200_2011}). 

With focus on the knowledge model, Sarker et al,. 2016 proposed a system that explains the classifier's outputs based on the background knowledge~\cite{Sarker2017}. The key tool of the system, called DL-Learner, works in parallel with the classifier and accepts the same data as input. Using the Suggested Upper Merged Ontology (SUMO)\footnote{http://www.adampease.org/OP/} as the symbolic knowledge model, the DL-Learner is also able to categorize the images by reasoning upon the objects together with the concepts defined in the ontology. The compatibility between the output of the DL-Learner and the classifier can be seen as a reliability support and at the same time as an interpretation of the classification process. 

Similarly, Icarte et al., 2017, introduced a general-purpose knowledge model called the ConceptNet Ontology~\cite{Icarte2017}. In this work, the integration of the symbolic model and a sentence-based image retrieval process based on deep learning is used to improve the performance of the learning process. The knowledge about different concepts, such as their affordances and their relations with other objects, is aligned with objects derived from the deep learning method.

The method of enriching the data by providing information as additional channels for training a CNN-based network has been done before. Liu et al., 2018 and Zhenyi et al., 2018, have explained how to augment the input data by adding two additional channels that represent the $i$ and $j$ coordinates in the image to obtain the location information~\cite{2018arXiv180703247L, wang2018location}. Our work uses information from a semantic referee as the augmented data instead of the location information.

Although in these works the role of symbolic knowledge represented by ontologies has been emphasized, they are limited in terms of the symbolic representation models. More specifically, the concepts and their relations in ontologies are simplified, limiting the richness of deliberation in an eventual reasoning process, especially for visual imagery data. 

Our approach can also be compared with Explanation-based learning (EBL)~\cite{ebl-learning} approaches. EBL refers to a form of machine learning method that is able to learn by generalizing examples where the features of the examples are formalized as domain theory. In EBL, the explanations, which consist of the features of the observation, are directly considered and generalized by the learner, whereas in our semantic based model, although the features of the misclassified regions are inferred from the ontology and send back to the classifier, they are not directly applied on the classification output; rather, they are only treated as a new set of data that is sent through the learning process.

\section{Method}
\label{sec:method}

\subsection{Overview of the approach}
\label{sec:approach}

\begin{figure*}[t]
\centering
\includegraphics[width=\textwidth, height=0.42\textheight]{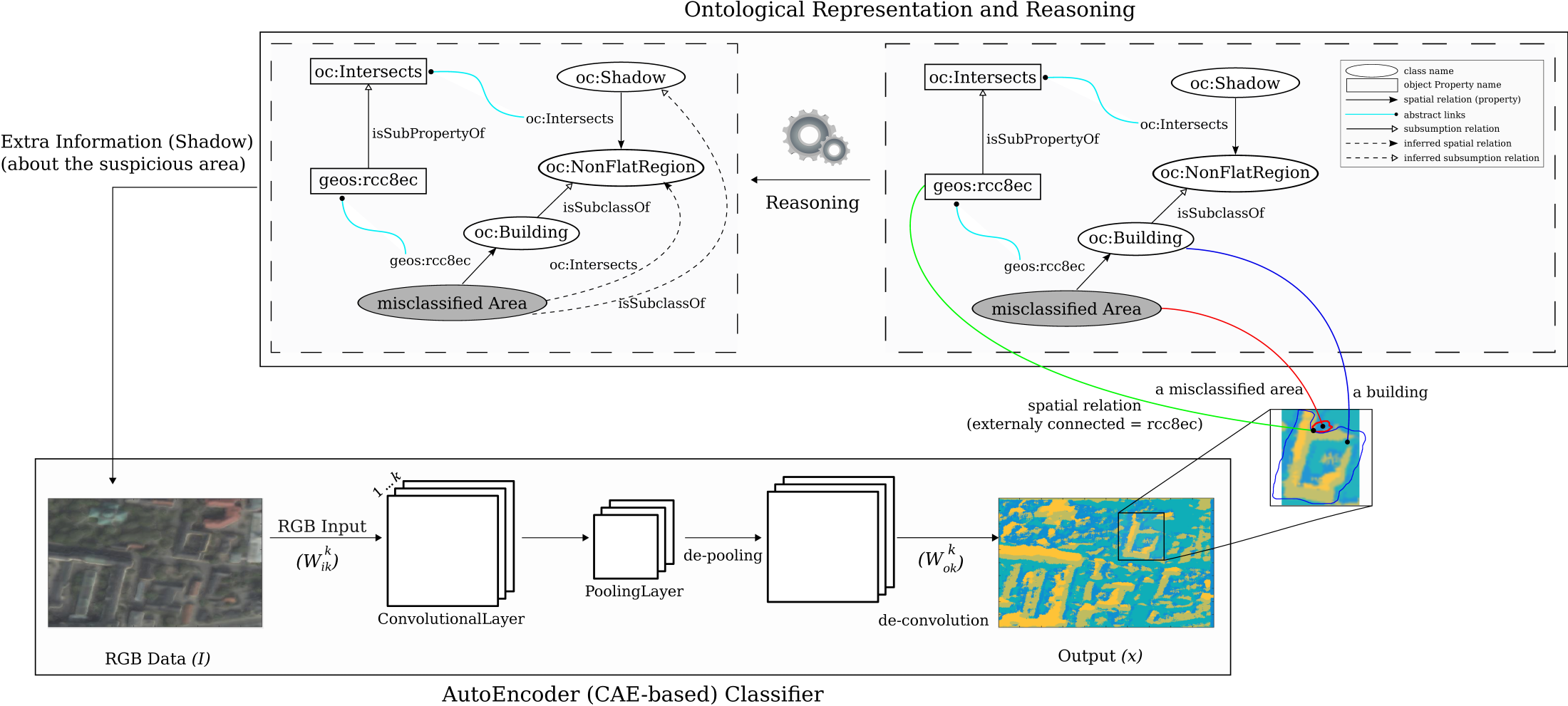}
\caption{Overview of applying a semantic referee (top layer) in the form of reasoning upon ontological knowledge to improve the semantic segmentation task of the convolutional encoder-decoder classifier (bottom layer). The semantic referee reasons about the mistakes made by the classifier based on ontological concepts and provides additional information back to the classifier that prevents the classifier from making the same misclassifications. The ontological knowledge and reasoning methods play the role of the semantic referee that makes sense of the errors from the classifier.}
\label{fig:flow-1}
\end{figure*}


An overview of our approach can be seen in Figure~\ref{fig:flow-1}, which shows the interaction between the classifier and the semantic referee. 

The classifier is a deep convolutional network with an encoder-decoder structure that provides the semantic segmentation of the input image (see Section~\ref{sec:dataclassification}). 

In order to deal with the misclassifications from the classifier, a semantic referee reasons about the errors that include the conceptualization of the misclassified regions based on their physical (e.g. geometrical) properties and aligns them with the available ontology to infer the best possible match for the error (see Section~\ref{sec:ontocity} and~\ref{sec:expmis}). 

The inferred concept related to the misclassified region is then given to the classifier as a referee providing information to be used within the learning process to prevent the classifier from making the same misclassifications. The additional information provided by the reasoner is represented as image channels with the same size as the RGB input and is then concatenated together with the original RGB channels as additional color channels. In this work, we use three additional channels from the reasoner that represents shadow estimation, elevation estimation, and other inconsistencies. This results in the classifier using data with 6 channels instead of the original 3 RGB channels (see Section~\ref{sec:improve}).

This process is then repeated until the classification accuracy on the validation data converges. During testing, the same procedure is performed using the same number of iterations that was used during training. 

\subsection{Data}
\label{sec:data}

\subsubsection{Stockholm and Boden}

\begin{figure}[t!]
\centering
\subfigure[Test data Boden]{\includegraphics[width=0.20\textwidth]{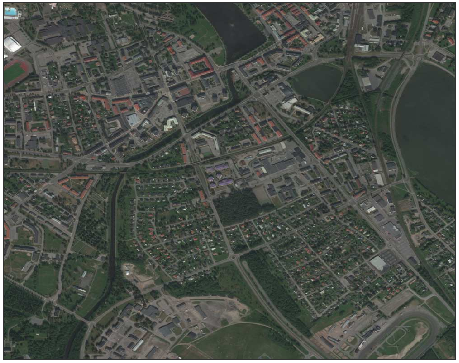} \label{fig:test_data_boden}} 
\subfigure[Train data Boden]{\includegraphics[width=0.20\textwidth]{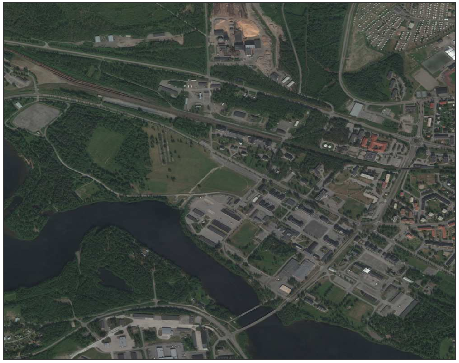} \label{fig:train_data_boden}} \\
\subfigure[Test data Stockholm]{\includegraphics[width=0.20\textwidth]{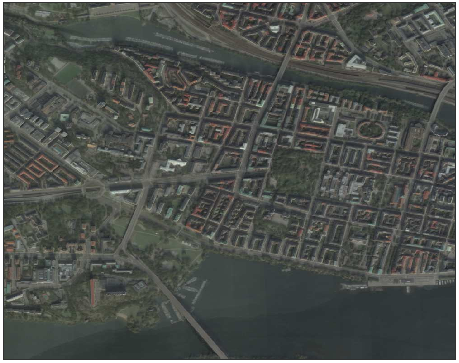} \label{fig:test_data_stockholm}} 
\subfigure[Train data Stockholm]{\includegraphics[width=0.20\textwidth]{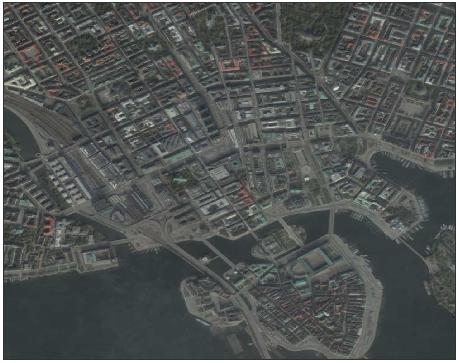} \label{fig:train_data_stockholm}} 
\caption{The data consists of RGB satellite images from two different cities Stockholm and Boden. The selected area size for both cities is $4000 \times 8000$ pixels with a pixel-resolution of $0.5$ meters and was divided into train and test sets with a $50$--$50$ split.}
\label{fig:data}
\end{figure}

The data used in this work consists of RGB satellite images from two different cities in Sweden, shown in Figure~\ref{fig:data}. The first city is Stockholm, which is the largest city and capital of Sweden, and the second city is a smaller city located in northern Sweden called Boden. The selected area size for both cities is $4000 \times 8000$ pixels with a pixel-resolution of $0.5$ meters and was divided into train and test sets with a $50$--$50$ split. The ground truth used for supervised training and evaluation has been provided by Lantm\"{a}teriet, the Swedish Mapping, Cadastral and Land Registration Authority.\footnote{\url{https://www.lantmateriet.se/}} 

The 5 categories that are used are \textit{vegetation}, \textit{road}, \textit{building}, \textit{water}, and \textit{railroad}. The class distribution for each city can be seen in Table~\ref{table:classdistributiontable}. Due to the large imbalance in the data set, the loss function uses median frequency class weighting. 
\begin{table*}[!h]
\centering
\begin{tabular}{c | c c c c c}
  \%              & Vegetation & Road & Building & Water & Railroad \\ \cline{1-6}
Stockholm (train) & 7.6        & 31.3 & 35.4     & 23.5  & 2.2 \\   
Stockholm (test)  & 18.2       & 36.9 & 19.7     & 22.4  & 2.8 \\
Boden (train)     & 63.0       & 19.7 & 4.9      & 11.0  & 1.3 \\   
Boden (test)      & 54.5       & 25.6 & 10.8     & 7.2   & 1.8 \\
   \end{tabular}
   \caption{Class distribution for the two cities used in this work. There is large difference in amount of vegetation, building, and water between the two cities. Both cities have a very small amount of railroads.}
   \label{table:classdistributiontable}
\end{table*}

\subsubsection{UC Merced Land Use Dataset}
\label{sec:dataucmerced}

For further evaluation, we use the publicly available UC Merced Land Use dataset~\cite{yang2010bag}, which consists of 21 land use classes with $~100$ images for each class with size $256 \times 256$ pixels and a pixel resolution of 1 foot. This data set is labeled by the land use for each image. The work by Shoa et al., 2018, has labeled each pixel in each image into 17 new classes~\cite{shao2018performance}. The new data set, called DLRSD, that is densely labeled can be used for remote sensing image retrieval (RSIR) and semantic segmentation. For this work, the 17 classes were merged into 8 new classes that were semantically similar in order to reduce the number of classes and resemble the already defined class-rules in the reasoner. The new merged classes, what classes they were merged from, and the prevalence of each new class are seen in Table~\ref{table:classdistributiontable_ucmerced}. The introduction of new classes might require some modification in the reasoner. In this work, we have added a rule about the size of the object in the reasoner since the three new classes \textit{airplane}, \textit{car}, and \textit{ship} are generally smaller than the other classes. We removed all images that have \textit{field} or \textit{tennis court} since they require a more complex class definition.

\begin{table*}[!h]
\centering
\begin{tabular}{c|c|c} \\ 
New class & Old classes & Prevalence [\%] \\ \cline{1-3}
Vegetation            & trees, \sout{field}, grass               & 28.6 \\
Non-vegetation ground & bare soil, sand, chaparral, \sout{court} & 17.6 \\
Pavement              & pavement, dock                           & 27.4 \\
Building              & building, mobile home, tank              & 13.7 \\
Water                 & water, sea                               & 7.9 \\
Airplane              & airplane                                 & 0.4 \\
Car                   & cars                                     & 2.9 \\
Ship                  & ship                                     & 1.6 \\
   \end{tabular}
   \caption{New merged classes and old classes from the DLRSD data set and class distribution on the new class. Any images containing the crossed out classes where omitted.}
   \label{table:classdistributiontable_ucmerced}
\end{table*}

\subsection{Data classification}
\label{sec:dataclassification}

\sloppy
A variation of a Convolutional Auto-encoder (CAE)~\cite{masci2011stacked} is used to perform the semantic segmentation of the satellite images where every pixel in the map is classified. The structure of the networks follows the model U-net~\cite{unet} and is created in \textit{MATLAB 2018a} with the function \textit{creatUnet} with patch size 256, 6 color channels (3 RGB channels + 3 channels as the feedback provided by the semantic referee (see Figure~\ref{fig:flow-1})), and 5 classes shown in Table~\ref{table:classdistributiontable}.

The U-net model consists of an encoder with 4 layers where each layer performs two convolutions with $64*L$ filters in the $L^{th}$ layer and a $2\times 2$ max-pooling operation; followed by two convolutions with $1024$ filters and $50\%$ dropout; and finally, a decoder with 4 layers that performs upsampling with a scaling factor of 2, depth concatenation with the output from the encoder at the same layer, and two convolutions. The number of filters in each layer $L$ in the decoder is $512/L$. Each convolution has a filter size of $3\times 3$ and is followed by a ReLu-activation~\cite{nair2010rectified}. The classification of each pixel is performed with a convolution with $k$ filters, where $k$ is the number of classes, with filter size $1\times 1$ followed by a \textit{softmax} activation function for the final per-pixel classification.

The model parameters are trained from scratch and were initialized with Xavier initialization~\cite{glorot2010understanding} and trained using the Adam optimization method~\cite{Kingma2014AdamAM} with initial learning rate $10^{-4}$ and minibatch size $20$ with early-stopping using a validation set that was randomly drawn from $10\%$ of the training data.

\subsection{OntoCity: the ontological knowledge model}
\label{sec:ontocity}

\begin{figure}[!ht]
\centering
\includegraphics[width=0.41\textwidth]{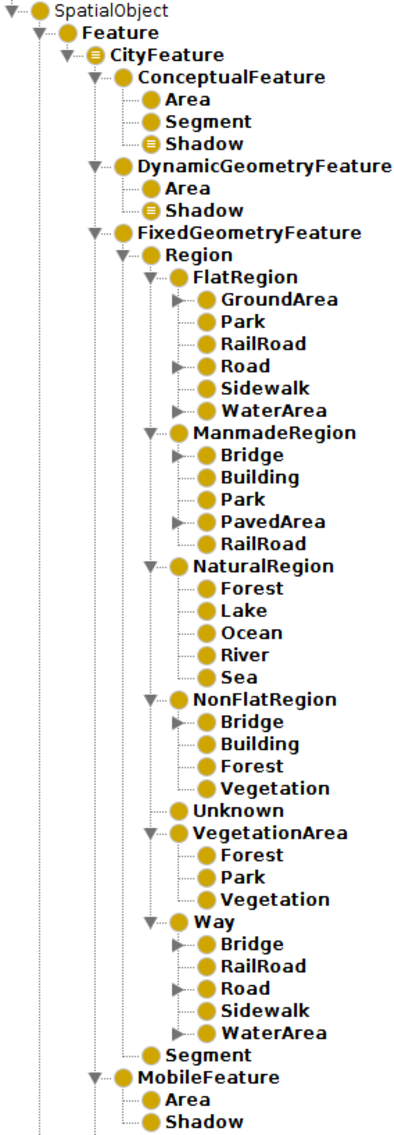}
\caption{A snapshot of the hierarchy of concepts in OntoCity. The city features are defined as the subclasses of the \texttt{geos:Feature} class defined in GeoSPARQL.}
\label{fig:ontocity}
\end{figure}

In our approach the improvement of data classification relies on an ontological reasoning process. The ontology that we have used as the knowledge model is called OntoCity\footnote{https://w3id.org/ontocity/ontocity.owl} and contains the domain knowledge about generic spatial constraints in outdoor environments. OntoCity  whose (part of) representational details explained by Alirezaie et al., 2017 ~\cite{DBLP:journals/sensors/AlirezaieKLKL17}, is an extension of the GeoSPARQL ontology, known as a standard vocabulary for geospatial data~\citep{DBLP:conf/rweb/KoubarakisKKNS12}. The main idea behind designing OntoCity was to develop a generalized knowledge model for representation of cities in terms of their structural, conceptual and physical aspects.
Figure~\ref{fig:ontocity} illustrates a Prot\'{e}g\'{e}~\cite{Musen:2015:PPL:2757001.2757003} snapshot of the hierarchy of concepts defined in OntoCity.

The class \texttt{\small{oc:CityFeature}} is one of the general classes defined in OntoCity, and is subsumed by the concept \texttt{\small{geos:Feature}} in GeoSPARQL. As can be seen, the name of a class contains a prefix that indicates the ontology to which it belongs. In the aforementioned classes, the two prefixes \texttt{\small{oc}} and \texttt{\small{geos}} refer to the URIs (Uniform Resource Identifiers) of the two ontologies OntoCity and GeoSPARQL, respectively. 

The class \texttt{\small{geos:Feature}}, which represents any spatial entity with some geometry, subsumes the class \texttt{\small{oc:CityFeature}} representing features in a city in the form of polygons that share at least one spatial relation with the remaining features. The axioms of OntoCity given in this paper are in description logic (DL)~\cite{Baader:2003:BDL:885746.885749}:

\begingroup
\fontsize{7.5pt}{7pt}\selectfont
\centering
\begin{flalign}
& \texttt{oc:CityFeature} \hspace{0.1cm}  \sqsubseteq \hspace{0.1cm} \texttt{geos:Feature} \hspace{0.1cm} \sqcap \\
\nonumber & \hspace{1cm} \exists \hspace{0.1cm} \texttt{geos:hasGeomtery}.\texttt{geos:Polygon} \hspace{0.1cm} \sqcap \\
\nonumber & \hspace{1cm} \exists \hspace{0.1cm} \texttt{oc:hasSpatialRelation}.\texttt{oc:CityFeature}
\end{flalign}
\endgroup

Spatial relations in OntoCity include the \mbox{RCC-8} (Region Connection Calculus) relations defined by Cohn et al., 1997 ~\cite{Cohn97qualitativespatial} and adopted by GeoSPARQL, with a bit of extension. The extension includes the definition of the relation \texttt{\small{oc:intersects}} that subsumes several RCC-8 relations including partially overlapping (\texttt{\small{geos:rcc8po}}) and externally connected (\texttt{\small{geos:rcc8ec}}). The spatial relation \texttt{\small{oc:intersects}} is used when we are only interested to know whether two given features intersect or not, regardless of the RCC-8 spatial relation between them.

Spatial relations are used in the form of spatial constraints to provide meaning to the city features. City features are categorized into several types defined as the  subclasses of \texttt{\small{oc:CityFeature}} in OntoCity. These categories include  \texttt{\small{oc:PhysicalFeature}} and \texttt{\small{oc:ConceptualFeature}}, which represent features with physical geometry (e.g. a landmark with an absolute elevation value measured from the sea floor), or conceptual geometry (e.g. a rectangular division in a city regardless of their landmarks), respectively. 

\begingroup
\fontsize{7pt}{6pt}\selectfont
\centering
\begin{flalign}
& \texttt{oc:PhysicalFeature} \hspace{0.1cm}  \sqsubseteq \hspace{0.1cm} \texttt{oc:CityFeature} \hspace{0.1cm} \sqcap \\
\nonumber & \hspace{0.8cm} \exists \hspace{0.1cm} \texttt{oc:hasAbsoluteElevationValue}.\texttt{xsd:double}\\
& \texttt{oc:ConceptualFeature} \hspace{0.1cm}  \sqsubseteq \hspace{0.1cm} \texttt{oc:CityFeature}
\end{flalign}
\endgroup

Furthermore, the two other classes \texttt{\small{oc:FixedGeometryFeature}} and  \texttt{\small{oc:DynamicGeometryFeature}} represent features whose geometries are fixed or dynamic (changing in time) respectively. Mobility is another property that categorizes the city features into mobile (\texttt{\small{oc:MobileFeature}}, e.g. a car), or stationary (\texttt{\small{oc:StationaryFeature}}, e.g. a building): 

\begingroup
\fontsize{7pt}{6pt}\selectfont
\centering
\begin{flalign}
 & \texttt{oc:FixedGeometryFeature} \hspace{0.1cm}  \sqsubseteq \hspace{0.1cm} \texttt{oc:CityFeature}\\
  & \texttt{oc:DynamicGeometryFeature} \hspace{0.1cm}  \sqsubseteq \hspace{0.1cm} \texttt{oc:CityFeature}\\
 & \texttt{oc:MobileFeature} \hspace{0.1cm}  \sqsubseteq \hspace{0.1cm} \texttt{oc:CityFeature}\\
 & \texttt{oc:StationaryFeature} \hspace{0.1cm}  \sqsubseteq \hspace{0.1cm} \texttt{oc:CityFeature}
\end{flalign}
\endgroup

As shown in Figure~\ref{fig:ontocity}, each of the subclasses of the class \texttt{\small{oc:CityFeature}} has its own taxonomy. For example, as shown in axiom~(\ref{axi:region}), the class \texttt{\small{oc:Region}} as a \textit{physical} feature with a \textit{fixed geometry}  that is also \textit{stationary} (i.e., non-mobile), represents a landmark that can per se be categorized into various types such as flat or non-flat, or likewise, into man-made or natural regions.

\begingroup
\fontsize{7pt}{6pt}\selectfont
\centering
\begin{flalign}
\label{axi:region}
 & \texttt{oc:Region} \sqsubseteq \hspace{0.1cm} \texttt{oc:PhysicalFeature} \hspace{0.1cm} \sqcap \hspace{0.1cm}\\
\nonumber & \hspace{0.7cm} \texttt{oc:StationaryFeature} \hspace{0.1cm} \sqcap\\
\nonumber & \hspace{0.7cm} \texttt{oc:FixedGeometryFeature}
\hspace{0.1cm}
\end{flalign}
\endgroup

\begingroup
\fontsize{7pt}{6pt}\selectfont
\centering
\begin{flalign}
&\texttt{oc:ManmadeRegion} \hspace{0.1cm} \sqsubseteq \hspace{0.1cm} \hspace{0.1cm} \texttt{oc:Region}
\end{flalign}
\endgroup

\begingroup
\fontsize{7pt}{6pt}\selectfont
\centering
\begin{flalign}
&\texttt{oc:NaturalRegion} \hspace{0.1cm} \sqsubseteq \hspace{0.1cm} \hspace{0.1cm} \texttt{oc:Region}
\end{flalign}
\endgroup

\begingroup
\fontsize{7pt}{6pt}\selectfont
\centering
\begin{flalign}
&\texttt{oc:FlatRegion} \hspace{0.1cm} \sqsubseteq \hspace{0.1cm} \hspace{0.1cm} \texttt{oc:Region}
\end{flalign}
\endgroup

\begingroup
\fontsize{7pt}{6pt}\selectfont
\centering
\begin{flalign}
\label{axi:nonflat}
&\texttt{oc:NonFlatRegion} \hspace{0.1cm} \sqsubseteq \hspace{0.1cm} \hspace{0.1cm} \texttt{oc:Region} \hspace{0.1cm} \sqcap\\
\nonumber & \hspace{0.7cm} \exists \hspace{0.1cm} \texttt{oc:hasRelativeElevationValue}.\texttt{xsd:double} \hspace{0.1cm} \sqcap\\
\nonumber & \hspace{0.7cm} \exists \hspace{0.1cm} \texttt{oc:intersects}.\texttt{oc:Shadow}
\end{flalign}
\endgroup

For each location in a city (or in general on the ground) there are two elevation values, namely absolute elevation and relative elevation. The absolute elevation is the value measured from the sea-level with height value zero, whereas the relative elevation value of a specific location indicates its relative height w.r.t the ground level and its vicinity. By a non-flat region we refer to landmarks of a city with a non-zero relative elevation value (see axiom~(\ref{axi:nonflat})). Due to its height, a non-flat region is also assumed to cast shadows. As shown in axiom ~(\ref{axi:shadow}), the concept of shadow has been also defined in OntoCity (\texttt{\small{oc:Shadow}}) due to its spatial relations with the other city features.

The texture of regions (i.e., landmarks) are defined as subclasses of the class \texttt{\small{oc:Region}}. It is worth mentioning that some of these region types are equivalent to the labels (i.e., classes listed in Table~\ref{table:classdistributiontable}) taken into account by the classifier to classify regions. These regions are defined as follows:

\begingroup
\fontsize{7.5pt}{7pt}\selectfont
\centering
\begin{flalign}
 & \texttt{oc:River} \hspace{0.1cm} \sqsubseteq \texttt{oc:WaterArea} \sqsubseteq \hspace{0.1cm} \texttt{oc:Region} \\
 & \texttt{oc:Road} \hspace{0.1cm} \sqsubseteq \texttt{oc:PavedArea} \sqsubseteq \hspace{0.1cm} \texttt{oc:ManmadeRegion} \\
 & \texttt{oc:Park} \hspace{0.1cm} \sqsubseteq \texttt{oc:VegetationArea} \sqsubseteq \hspace{0.1cm} \texttt{oc:Region} \\
 & \texttt{oc:Building} \sqsubseteq \hspace{0.1cm} \texttt{oc:ManmadeRegion} \hspace{0.1cm} \sqcap \\
\nonumber & \hspace{0.7cm} \hspace{0.1cm} \texttt{oc:NonFlatRegion}
\end{flalign}
\endgroup

Each  class of these region types can have more than one superclass. For instance, a railroads is defined as a flat (not high) man-made region which is used as a way (i.e., route) in a city. Given this definition, the main three constraints in the definition of the concept \texttt{\small{oc:RailRoad}} are in the form of three subsumption relationship with the concepts \texttt{\small{oc:ManmadeRegion}}, \texttt{\small{oc:FlatRegion}} and \texttt{\small{oc:Way}} as follows:

\begingroup
\fontsize{7.5pt}{7pt}\selectfont
\centering
\begin{flalign}
 & \texttt{oc:RailRoad} \hspace{0.1cm} \sqsubseteq \texttt{oc:ManmadeRegion} \sqsubseteq \hspace{0.1cm} \texttt{oc:Region} \\
 & \texttt{oc:RailRoad} \hspace{0.1cm} \sqsubseteq \texttt{oc:FlatRegion} \\
 & \texttt{oc:RailRoad} \hspace{0.1cm} \sqsubseteq \texttt{oc:Way} \sqsubseteq \hspace{0.1cm} \texttt{oc:Region}
\end{flalign}
\endgroup

The RCC-8 relations are used to complete the definition of the region types or describe more specific features (e.g. bridges, shadows, shores) whose definitions rely on their spatial relations with their vicinity. For instance, as defined in axiom~(\ref{axi:railroad}), a railroad in a city is expected to intersect (more specifically externally connect) with at least one but only with features as either a vegetation area, dirt or ground.

\begingroup
\fontsize{7.5pt}{7pt}\selectfont
\centering
\begin{flalign}
\label{axi:railroad}
 & \texttt{oc:RailRoad}  \hspace{0.1cm} \sqsubseteq \hspace{0.1cm} \forall \hspace{0.1cm} \texttt{oc:intersects}.(\texttt{oc:Dirt} \hspace{0.1cm} \sqcup\\
 \nonumber & \hspace{0.7cm} \texttt{oc:VegetationArea} \hspace{0.1cm} \sqcup \hspace{0.1cm} \texttt{oc:Ground})\hspace{0.1cm} \sqcap \\
\nonumber & \hspace{0.7cm} \exists \hspace{0.1cm} \texttt{geos:rcc8ec}.\texttt{oc:Region} \hspace{0.1cm} 
\end{flalign}
\endgroup

Likewise, a bridge is a man-made non-flat region that is partially overlapping (referring to the RCC-8 relation \texttt{\small{geos:rcc8po}}) at least one other region, the texture of which identifies the bridge type. If the region is a water-area then the overlapping bridge is a water bridge, or if the region is a street, then the bridge is categorized as a street or a pedestrian bridge:

\begingroup
\fontsize{7.5pt}{7pt}\selectfont
\centering
\begin{flalign}
 & \texttt{oc:Bridge} \hspace{0.1cm} \sqsubseteq \texttt{oc:ManmadeRegion} \hspace{0.1cm} \sqcap \\
\nonumber & \hspace{0.7cm} \hspace{0.1cm} \texttt{oc:NonFlatRegion} \hspace{0.1cm} \sqcap\\
\nonumber & \hspace{0.7cm} \exists \hspace{0.1cm} \texttt{geos:rcc8po}.\texttt{oc:Region}
\end{flalign}
\endgroup

As one of the non-physical (conceptual) features defined in OntoCity, we can refer to the concept of shadow as a spatial feature with a dynamic and also mobile geometry (i.e., changing depending on the time of the day). Although the exact shape of shadows and their exact positions depend on many quantitative parameters including the position of the source light and the height value of the casting objects, it is still possible to  qualitatively describe shadows in the ontology. The definition of the concept shadow in OntoCity is more precise because it also contains a spatial constraints saying that, for the concept to be a shadow, it needs to intersect (\texttt{\small{oc:intersects}}) with at least one non-flat region which casts the shadow:

\begingroup
\fontsize{7.5pt}{7pt}\selectfont
\centering
\begin{flalign}
\label{axi:shadow}
 & \texttt{oc:Shadow} \hspace{0.1cm} \sqsubseteq \hspace{0.1cm}  \texttt{oc:ConceptualFeature} \hspace{0.1cm} \sqcap \\
\nonumber & \hspace{0.7cm} \texttt{oc:DynamicGeometryFeature} \hspace{0.1cm} \sqcap \\
\nonumber & \hspace{0.7cm}\hspace{0.1cm} \texttt{oc:MobileFeature} \hspace{0.1cm} \sqcap\\
\nonumber & \hspace{0.7cm} \exists \hspace{0.1cm} \texttt{oc:intersects}.\texttt{oc:NonFlatRegion}
\end{flalign}
\endgroup

\subsubsection{Specialization of OntoCity}
\label{sec:ontostockholm}
The OntoCity axioms mentioned in the previous sections are a subset of general knowledge that always holds regardless of the city under study (e.g. ``\textit{Water bridges cross water areas}''). However, depending on the case study, the background knowledge might be specialized to represent features belonging to a specific environment (e.g. ``\textit{in the given region there is no building connected to water areas}'').

The areas under our study, as shown in Section~\ref{sec:data}, comprise the central part of Stockholm and also another small city Boden in north of Sweden. The following spatial constraints are valid for both of these cities and that is why they have been added to the version of OntoCity used in our case:

\begin{enumerate}
\item Buildings are directly connected to at least a road or a vegetation area  (referring to the connected relation in RCC8: \texttt{\small{geos:rcc8ec}} relation) 
\item Buildings do not intersect with railroads (referring to the negation of the \texttt{\small{oc:intersects}} relation) 
\item Buildings are not directly connected to water-area (referring to the  negation of externally connected relation in RCC8: \texttt{\small{geos:rcc8ec}}).
\item Buildings are not directly connected to rail roads (referring to the  negation of externally connected relation in RCC8: \texttt{\small{geos:rcc8ec}}).
\item Buildings are not contained by roads (referring to the negation of tangential proper part relation in RCC-8: \texttt{\small{geos:rcc8tpp}}).
\item Buildings do not contain roads (referring to the negation of tangential proper part inverse relation in RCC-8: \texttt{\small{geos:rcc8tppi}}).
\item Railroads are not directly connected to water-area (referring to the negation of the \texttt{\small{oc:intersects}} relation).
\end{enumerate}

The following axiom shows the DL definition of the class  \texttt{\small{oc:StockholmBuilding}} as the subclass of the class \texttt{\small{oc:Building}}:

\begingroup
\fontsize{7.5pt}{7pt}\selectfont
\centering
\begin{flalign}
 & \texttt{oc:StockholmBuilding} \hspace{0.1cm} \sqsubseteq \hspace{0.1cm}  \texttt{oc:Building} \hspace{0.1cm} \sqcap \\
\nonumber & \hspace{0.7cm} \hspace{0.1cm} \exists \hspace{0.1cm} \hspace{0.1cm} \texttt{geos:rcc8ec}.(\texttt{oc:VegetationArea} \sqcup	 \texttt{oc:Road}) \hspace{0.1cm} \sqcap \\
\nonumber & \hspace{0.7cm} \not \exists \hspace{0.1cm} \texttt{oc:intersects}.\texttt{oc:RailRoad} \hspace{0.1cm} \sqcap \\
\nonumber & \hspace{0.58cm} \not \exists \hspace{0.1cm} \texttt{geos:rcc8ec}.\texttt{oc:Waterarea} \hspace{0.1cm} \sqcap \\
\nonumber & \hspace{0.58cm} \not \exists \hspace{0.1cm} \texttt{geos:rcc8tpp}.\texttt{oc:Road} \hspace{0.1cm} \sqcap \\
\nonumber & \hspace{0.58cm} \not \exists \hspace{0.1cm} \texttt{geos:rcc8tppi}.\texttt{oc:Road}
\end{flalign}
\endgroup

The spatial constraints  used in the definition of classes are considered by a reasoner in order to discard the impossible labels (region types) for a region based on its neighborhood.

\subsection{Semantic Augmentation of Errors}
\label{sec:expmis}
The size of each patch of data (either testing or training data for Boden or Stockholm) is $4000 \times 4000$ pixels. A segmentation is performed on each patch using the MATLAB function \textit{superpixels}, which resulted in around $30000$ regions for each patch. This segmentation is applied to reduce the computational complexity of the reasoning process by only considering the spatial relations of each region with its vicinity i.e., with regions located in a same segment.

The output of the classifier is in the form of labeled pixels. Given a set of pixels carrying the same class label, the semantic referee, within a geometrical process\footnote{The  extraction process is done using the predefined MATLAB function, \textit{bwboundaries}, used to trace region boundaries in binary images.}, first extracts the boundary of these pixels to form a polygon with the region type equivalent to the class label. Each polygon (region) is also assigned with a probability of classification certainty. The regions with low certainty are suspected to be misclassified ones and should be prioritized for inspection by the reasoner. 

To get the discrete classification of each pixel an argmax follows the softmax layer (that outputs class probabilities). To calculate the predicted class for each region, the class that has the highest mode after the argmax of the class probabilities of each pixel in the region is selected. Then to calculate the classification certainty of the region, we take the average class probabilities of the predicted region class for each pixel in the region.

Given the output of the classification together with ontological knowledge about city features, the  reasoner as a semantic referee semantically augments the errors based on the content of the ontology. The process is composed of several steps which are in brief captured in Algorithm~\ref{algo:semexp}. 

\begingroup
\fontsize{6pt}{6pt}\selectfont	
\begin{algorithm} 
 \small
\caption{Error Semantic Augmentation}  
\label{algo:semexp} 
\begin{algorithmic}[1]
\Require $W = empty, S, m, R $ \\
\% W: A hash-map, empty in the beginning\\
\% S: The given list of rectangular segments \\
\% P: The given list of misclassified regions \\
\% R: The given list of classified regions
\ForEach {$s \in S $}
\State $ P_{s}\gets \operatorname{getRegionsInSegment}(P, s)$ 
\State $ R_{s}\gets \operatorname{getRegionsInSegment}(R, s)$ 
\ForEach {$r \in  R_{s} $}
\State $ t \gets \operatorname{getRegionType}(r)$
\ForEach {$p \in  P_{s} $}
\State $ q\gets \operatorname{calculateRCC}(p, r)$ 
\State $ W.\operatorname{add}(< q, t>) $ 
\EndFor
\EndFor
\EndFor
\State $<Q, T> \gets \operatorname{getHighFrequentSpatialRelations}(W)$ 
\State $ C\gets \operatorname{queryOntology}(Q, T)$
\State $ augmentation \gets \operatorname{getSemantics}(C)$
\end{algorithmic}
\end{algorithm}
\endgroup

The algorithm accepts as input the list of segments ($S$) and the list of both classified ($R$) and misclassified ($P$) regions in the form of polygons. For each segment, the algorithm extracts all the classified ($R_{s}$) and misclassified regions ($P_{s}$) that belong to the segment.

Given the two lists of polygons $R_{s}$ and $P_{s}$, the algorithm calculates all the possible (RCC-8) qualitative spatial relations between any pairs of $(p, r)$ where $p \in P_{s}$ is a misclassified region and $r \in R_{s}$ is a classified region in its vicinity. 

For each pair $(p, r)$, the algorithm calculates the spatial relation \textit{q} between \textit{p} and \textit{r} and also keeps the type of the region $r$ named as \textit{t}. All the calculates pairs $<q, t>$ are added to the hash-map structure \textit{W}. The hash-map \textit{W} will at the end contain all the spatial relations that exist between the misclassified regions for each specific region type (see lines 5-15). In other words, \textit{W} is defined to contain the geometrical characteristics of the misclassified regions.

To find a general description indicating why the classifier has been confused, the characteristics of the errors are generalized based on their frequency. If we assume that the pair $<Q, T>$ (see line 16) represents the most observed spatial relation \textit{Q} between the misclassified regions and a specific region type \textit{T}, then this pair can be generalized and counted as a representative feature of the misclassified regions. Given the representative pair $<Q, T>$, the algorithm queries OntoCity to find all the spatial features that are at least in one \textit{Q} relation with the region type \textit{T}. The DL expression of the query is: $\exists~T.Q$.

By applying the ontological reasoner the query can also be further generalized from type \textit{T} to its super-classes in OntoCity (see line 17). The concept (\textit{C}) as a spatial feature (\textit{C} $ \sqsubseteq $ \texttt{oc:CityFeature}) inferred by the reasoner, is considered as the semantic augmentation for the misclassified regions that are in the given spatial relations with the given region type.

The computational complexity of the algorithm has the order of magnitude $O(k \times m \times n)$, where k is the number of segments, m is the number of classified and m is the average number of misclassified regions in the segment.

\subsection{Feedback from the reasoner to the classifier}
\label{sec:improve}

In this work, the reasoner will provide the feedback to the classifier in the form of additional information that will be augmented to the original RGB training data. The additional information is represented as an image with the same size as the input image that describes a certain property of a concept for each pixel of the original data. We have selected the following three concepts that the reasoner should give feedback about: shadow estimation, height estimation, and uncertainty information. This means that the input to the classifier will have 6 color channels (3 RGB channels + 1 shadow estimation channel + 1 height estimation channel + 1 uncertainty information channel) instead of the original 3 RGB channels. The classifier is then re-trained on this new input and provides a new semantic segmentation for the reasoner to reason about. The three concepts are described in more detail below.

\begin{description}

\item[Shadow] The first channel describes the presence of shadow, which is one major cause behind many of the misclassifications. There is a fair amount of research work with the focus on shadow detection in the fields of computer vision and pattern recognition~\cite{Sanin:2012:SDS:2107787.2108041}. In order to report the concept of shadow back to the classifier, we first need to localize them on the map. Although neither in OntoCity nor in other available ontologies is there any formal representation to calculate the location of shadows, this explanation as a semantic referee provides a significant insight for us to develop the reasoner to localize the shadows. The values for this channel are -1 (not shadow), 0 (no opinion), 1 (shadow).

\item[Elevation] Another property that has an influence on the classification and might be a cause for the misclassifications is elevation (second channel). Since elevation difference of regions is one of the main parameters in casting shadows, we have assigned the relative elevation value for each region as the average of its pixels' elevation values. Given the elevation value together with the type and the spatial relations of regions in the neighborhood of each misclassified region, the reasoner is able to localize the shadows as the group of pixels of the misclassified region with the lowest elevation value with respect to the elevation values of the regions intersecting with the misclassified region. The values for this channel are -1 (uncertain), 0 (low height), 1 (medium height), and 2 (high height). 

\item[Uncertainty] Finally, the third channel of data is dedicated to the pixels of those uncertain regions whose spatial relations with their neighborhood were found inconsistent w.r.t OntoCity's constraints. Furthermore, the ontological reasoner finds many uncertain areas whose spatial relations with their neighborhood were inconsistent and violating the constraints defined in OntoCity. The values for this channel are 0 (no opinion) and 1 (uncertain).

\end{description}

\section{Empirical Evaluation}
\label{sec:result}



\subsection{Error Characterization}
\label{sec:result-error-characterization}

Since the ground truth is available for our data, it is possible to calculate the certainty of misclassified regions. In this work, we use the classification certainty and select all the regions whose classification certainty is less than 70\% and consider them as (likely) misclassified regions. Given both the classified regions and the misclassified regions, as explained in Section~\ref{sec:expmis}, the reasoner is able to conceptualize the errors. The conceptualization process is based on extracting the spatial relations of the misclassified regions with their segmented neighborhood. This step has been implemented using the open-source JTS Topology Suite\footnote{https://github.com/locationtech/jts}, whose summary of results for Stockholm and Boden are shown in Table~\ref{tab:spatial-relation-stockholm} and Table~\ref{tab:spatial-relation-boden}. Each cell of the table represents number of misclassified regions that are in a spatial relation (given in the column header) with all the regions with a specific type (given in the row header).

Given the Stockholm test data classification outputs, the reasoner, in order to find a representative feature of the misclassified regions, considers the pair $<Q,T>$ as the most observed spatial relation \textit{Q} between the misclassified regions and a specific region type \textit{T}. Table~\ref{tab:spatial-relation-stockholm} shows the results of the first round of the classification. As we can see, the most observed spatial relations that involves $136$ misclassified regions is the pair $<Q$=~\texttt{\small{geos:rcc8ec}},~$T$=~\texttt{\small{oc:Building}}$>$.

\begingroup
\fontsize{5pt}{5pt}\selectfont	
\begin{table}[!ht]
\small
\centering
\begin{tabular}{|l|*{2}{c|}}\hline                                                                                        
\backslashbox{Type (t)}{Relation (q)}
&\makebox[4em]{\texttt{ec}}&\makebox[4em]{\texttt{po}}\\\hline\hline
\texttt{\small{oc:Building}} &\textbf{136}& 3 \\\hline
\texttt{\small{oc:Road}} &59&0\\\hline
\texttt{\small{oc:Water}} &11&0\\\hline
\end{tabular}
\caption{Summary of the inconsistent spatial features of errors in classification of \textbf{Stockholm test data}. Each cell value represents the number of misclassified regions involved in the given spatial relations with the given region type, where \texttt{ec} and \texttt{po} refer to the RCC-8 relations \textit{externally connected} and \textit{partially overlapping}, respectively.}
\label{tab:spatial-relation-stockholm}
\end{table}
\endgroup

\begingroup
\fontsize{5pt}{5pt}\selectfont	
\begin{table}[!ht]
\small
\centering
\begin{tabular}{|l|*{2}{c|}}\hline                                                                                         
\backslashbox{Type (t)}{Relation (q)}
&\makebox[4em]{\texttt{ec}}&\makebox[4em]{\texttt{nttpi}}\\\hline\hline
\texttt{\small{oc:Building}} &\textbf{164}&\textbf{118}\\\hline
\texttt{\small{oc:RailRoad}} &\textbf{93}& 0 \\\hline
\end{tabular}
\caption{Summary of the inconsistent spatial features in classification of \textbf{Boden test data}. Each cell value represents the number of misclassified regions involved in the given spatial relations with the given region type, where \texttt{ec} and \texttt{nttpi} refer to the RCC-8 relations \textit{externally connected} and \textit{non-tangential proper part inverse} (i.e., ogc:contains), respectively.}
\label{tab:spatial-relation-boden}
\end{table}
\endgroup

Given the pair \mbox{$<Q,T>$}, the reasoner queries the ontology with spatial constraints. The ontological reasoner that we have used in this work is the extended version of the reasoner Pellet, as an open-source Java based OWL 2 ontological reasoner~\cite{Sirin:2007:PPO:1265608.1265744}. The extension is in terms of filtering concepts based on their spatial constraints.

The Description Logic (DL) syntax of the query given to the reasoner is $\exists$~\texttt{\small{geos:rcc8ec}}.\texttt{\small{oc:Building}}  interpreted as ``\textit{all the entities that are at least in one} \texttt{\small{geos:rcc8ec}} \textit{relation with the region type} \texttt{\small{oc:Building}}''. The ontological reasoner results in a hierarchically linked concepts in the ontology from the most generalized to the most specialized (direct superclass) concepts satisfying the constraint given in the query. The satisfactory concept is explained as ``\textit{a mobile conceptual feature with a dynamic geometry}'' or more specifically a \texttt{\small{oc:shadow}} (as a direct answer of the query). In OntoCity, the concept shadow is defined based on the spatial constraint: $\exists \hspace{0.1cm} \texttt{\small{oc:intersects}}.\texttt{\small{oc:NonFlatRegion}}$, which is found by the reasoner as the generalization of the query $\exists\hspace{0.1cm}\texttt{\small{geos:rcc8ec}}.\texttt{\small{oc:Building}}$ (where $\texttt{\small{geos:rcc8ec}}~\sqsubseteq~\texttt{\small{oc:intersects}}$ and  $\texttt{\small{oc:Building}}~\sqsubseteq~\texttt{\small{oc:NonFlatRegion}}$) (see Figure~\ref{fig:flow-1}, top layer).

\begin{figure}[!ht]
\centering
\includegraphics[width=0.5\textwidth]{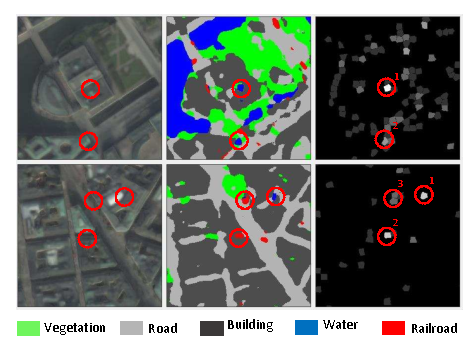}
\caption{Two examples of the Stockholm test set classification output along with their input RGB image, classified segmentation and the misclassification. The misclassified regions marked with numbers are in spatial relations with buildings, roads, vegetation, etc. The ontological reasoner can  augment the misclassification with the label shadow.}
\label{fig:results_noavg}
\end{figure}

Figure~\ref{fig:results_noavg} illustrates two samples taken from Stockholm test set classification output where the misclassified regions are marked in red. At the first row, the areas marked with number $1$ and $2$ are misclassified as water. As the RGB image on the left shows, the misclassified regions (in red) are (externally) connected to buildings that cast shadows. At the second row, the area marked with number $1$ is likewise misclassified as water. This area is again (externally) connected to a building. This area is also located between (i.e., connected with) at least two disconnected regions labeled as roads that are disconnected at the shadow area. This combination can explain the second most observed relation listed in Table~\ref{tab:spatial-relation-stockholm}, between the misclassified regions and the region type \texttt{\small{oc:Road}}. Assuming that buildings are often located close to roads (or streets), their shadow is likely casted on some parts of the roads. Therefore, a road instead of being recognized as a single road, is segmented into several roads disconnected at the shadow areas due to the change in their colors. Errors caused by shadows are not always labeled as water. Again in the second row, the areas marked with number 2 and 3 are also connected to buildings and roads, but are misclassified as railroads again due to the fact that the darkness of the shadow at this location is similar to the captured color of railroads in the image data.

Unlike Stockholm, in classification of Boden test data, most of the extracted spatial relations between misclassified regions and their vicinity were found inconsistent w.r.t. the constraints defined in OntoCity  (see Section~\ref{sec:ontostockholm}). As shown in Table~\ref{tab:spatial-relation-boden}, 93 rail roads were connected to other (misclassified) regions (e.g. buildings and water areas), a fact which according to OntoCity is inconsistent. Likewise, 164 buildings are connected to other regions, 67 cases out of which were again according to OntoCity inconsistent, and the remaining 97 cases (i.e., the consistent ones) were inferred as shadows. Moreover, 118 misclassified regions (mainly roads) were spatially contained in buildings. However, the ontological reasoner found them inconsistent as according to OntoCity roads cannot be contained (surrounded) by buildings.

\subsection{Classification accuracy results}
\label{sec:result-classification}

The following section presents the classification results on Stockholm, Boden, and UC Merced Land Use. The hardware that was used to train the classifiers for was a i7-8700K CPU @ 3.70Ghz with a GeForce GTX 1070 GPU. The time to train each classifier was around 3 hours for all data sets.

\subsubsection{Stockholm and Boden}
\label{sec:result-classification-stock-boden}

Two separate classifiers were trained on the training data for each of the two cities used in this work. Each classifier is then applied to the test data for both cities. The classifiers were first trained using a depth concatenation of the RGB channels and three channels that represent the estimations for elevation, shadow, and uncertain areas from the reasoner respectively. The additional channels are set to $0$ for the first round of training of the classifiers. On the subsequent iterations, the classifiers are then re-trained with the feedback from the reasoner in the form of adding information to the three additional channels. The process is repeated until the validation accuracy has converged. 

\begin{figure}[t!]
\centering
\subfigure[Trained on Stockholm, used on Stockholm test data.]{\includegraphics[width=0.20\textwidth]{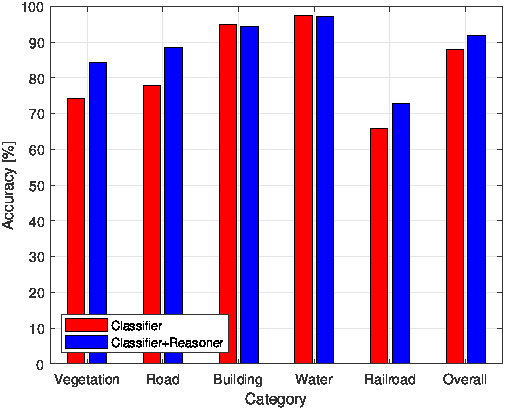} \label{fig:results1}} 
\subfigure[Trained on Stockholm, used on Boden test data]{\includegraphics[width=0.20\textwidth]{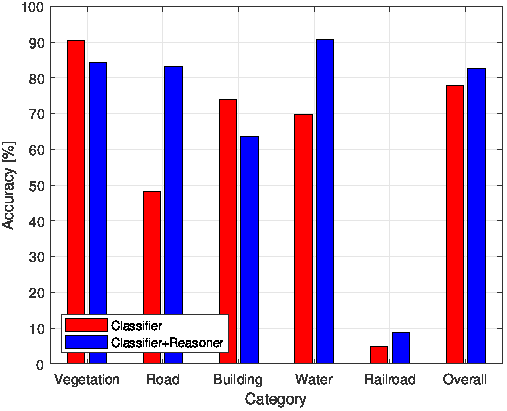} \label{fig:results2}} \\
\subfigure[Trained on Boden, used on Boden test data]{\includegraphics[width=0.20\textwidth]{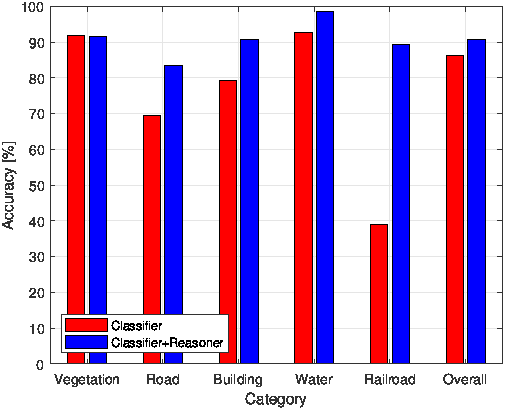} \label{fig:results3}} 
\subfigure[Trained on Boden, used on Stockholm test data]{\includegraphics[width=0.20\textwidth]{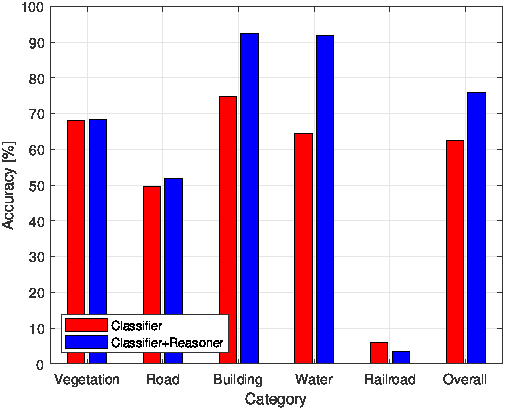} \label{fig:results4}} 
\caption{Classification results on the test data for both cities with two classifiers. The classifiers are trained separately on the training data for both cities before and after using the reasoner.}
\label{fig:results4figs}
\end{figure}

The per-class and overall classification accuracy on the test sets for both classifiers before and after the classifiers have been re-trained with the additional information from the reasoner and can be seen in Figure~\ref{fig:results4figs}. The overall accuracy is increased for all combinations of classifier and test data and almost all classes individually. The accuracy on the test data is higher if the classifier was trained on the same city and is decreased if the classifier was trained on another city. The class with the lowest accuracy when the classifier was trained on another city is railroad. The reason for this can be seen by observing that the railroads have different structure and surroundings between the two cities. The reasoner improved the results significantly for the test data on Boden with a classifier trained on the same city, see Figure~\ref{fig:results3}. 

\begin{figure*}[ht]
 \centering
\includegraphics[width=0.9\textwidth, height = 0.5\textheight]{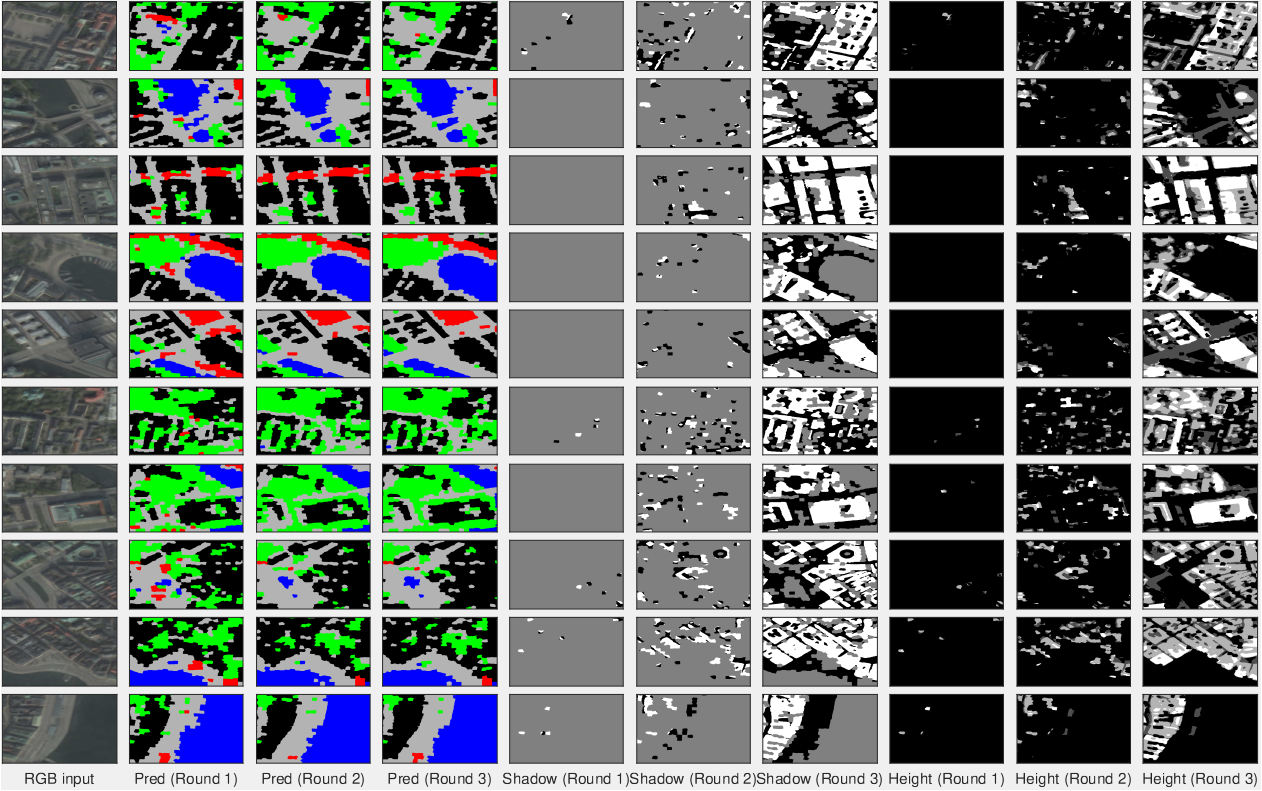}
\caption{RGB input (column 1), predictions from classifier for three rounds (column 2-4, green=vegetation, gray=road, black=building, blue=water, red=railroad), shadow estimations from reasoner for three rounds (column 5-7, gray=undefined, white=not shadow, black=shadow), and height estimations from reasoner for three round (column 8-10, black=low object, white=tall object).}
  \label{fig:PSEfig}
\end{figure*}

Some examples of the RGB inputs, predictions, shadow and height estimations for three rounds for Stockholm can be seen in Figure~\ref{fig:PSEfig}. The first round of training of the classifier results in a high number of misclassifications (column 2). When the reasoner has provided shadow estimation (column 5) and elevation information (column 8), the classifier is re-trained and gives an improved classification (column 3). The process is repeated until the validation accuracy has converged and most of the misclassifications have been corrected (column 4).

The confusion matrix for the last round on both test sets for a classifier that was trained on the Stockholm train data is given in Table~\ref{table:confmattable_stockholm}. The most difficult class to classify is the class \textit{railroad} and the largest confusion is between \textit{roads} and \textit{railroad}. The semantic referee improved the most for the class \textit{road} ($+20\%$) at the cost of introducing more confusion between the second largest confusion, which was between \textit{vegetation} and \textit{roads}. The reason behind the confusions regarding the class \textit{vegetation} can be related to their wide range of elevation values. The label \textit{vegetation} is not precise enough as it includes trees, lawns, parks, grass on the map and hence their elevation values are not informative enough to help the classifier to differentiate them from roads. The accuracy for \textit{buildings} is already high due to the large amount of buildings in Stockholm and the use of a reasoner did not improve the results for this class. The accuracy for \textit{water} is also high due to the large amount of water in Stockholm and was improved with the reasoner. One possible reason for this is that the water in the training set contains deeper water that has a darker color compared to the shallow water and channels that are present in Boden. 
\begin{table*}[!h]
\centering
\begin{tabular}{c c|c c c c c}
     & & \multicolumn{5}{c}{Predicted label} \\
     & \% & Vegetation & Road & Building & Water & Railroad \\ \cline{1-7}		 
     \multirow{5}{*}{\rotatebox{90}{Actual label}} 
& Vegetation & \textbf{84.2 (+4.55)} & 12.9 (-8.26)          & 2.3  (+4.13)           & 0.39 (-0.28)          & 0.22 (-0.14) \\
& Road       & 7.4 (+10.9)           & \textbf{86.4 (-20.0)} & 5.8  (+8.23)           & 0.31 (+0.30)          & 0.11 (-0.63) \\
& Building   & 1.2  (+0.61)          & 8.1 (-2.44)           & \textbf{90.6 (+1.86)}  & 0.12 (-0.08)          & 0.03 (-0.05) \\
& Water      & 2.3 (+2.85)           & 2.4 (+3.41)           & 0.01 (+0.31)           & \textbf{95.3 (-6.59)} & 0.00 (-0.01) \\
& Railroad   & 11.0 (+13.8)          & 37.3 (-15.8)          & 2.6 (+8.0)             & 0.27 (-0.17)          & \textbf{48.8 (-5.8)} \\
   \end{tabular}
   \caption{Confusion matrix [\%] for the test set for the classifier that was trained on Stockholm with the use of the reasoner. The numbers in parenthesis show how the result would change compared to a classifier that did not use a reasoner.}
   \label{table:confmattable_stockholm}
\end{table*}

When the classifier was trained on Boden, which has a smaller amount of training data for buildings and water, but more vegetation than Stockholm, the use of a reasoner improved the accuracy of buildings by $16.9\%$ and water by $20.5\%$ and achieved comparable results with the Stockholm classifier, see Table~\ref{table:confmattable_boden}. This illustrates that one of the strengths of using a reasoner is to compensate for classes with a small amount of training data. The reasoner also removes a large amount of previous confusion between \textit{buildings} and the two classes \textit{water} and \textit{railroad}. Similarly, Boden has a large amount of \textit{vegetation} and therefore already achieves a high accuracy on this class, and while the reasoner improves the results for all classes, it does not further improve the results for vegetation.
\begin{table*}[!h]
\centering
\begin{tabular}{c c|c c c c c}
     & & \multicolumn{5}{c}{Predicted label} \\
     & \% & Vegetation & Road & Building & Water & Railroad \\ \cline{1-7}		 
     \multirow{5}{*}{\rotatebox{90}{Actual label}} 
& Vegetation &   \textbf{89.2 (+0.16)} & 7.45 (-0.86)          & 2.70 (+0.61)          & 0.39 (-0.10)          & 0.24 (+0.19) \\
& Road       &   11.3 (+4.95)          & \textbf{64.1 (-6.84)} & 24.2 (+1.73)          & 0.31 (-0.20)          & 0.07 (+0.36) \\
& Building   &  4.0  (+4.03)           & 3.62 (+11.9)          & \textbf{92.3 (-16.9)} & 0.0 (+0.17)           & 0.0 (+0.85) \\
& Water      &  1.25 (+7.53)           & 4.54 (-3.58)          & 0.14 (+16.6)          & \textbf{94.1 (-20.5)} & 0.0 (+0.01) \\
& Railroad   &  10.4 (+8.28)           & 46.8 (-1.25)          & 6.92 (+10.2)          & 0.04 (+0.18)          & \textbf{35.8 (-17.4)} \\
   \end{tabular}
   \caption{Confusion matrix [\%] on the test set for the classifier that was trained on Boden with the use of the reasoner. The numbers in parenthesis show how the result would change compared to a classifier that did not use a reasoner.}
   \label{table:confmattable_boden}
\end{table*}

\subsubsection{UC Merced Land Use dataset}
\label{sec:result-classification-ucmerced}

A new classifier is trained on the UC Merced Land Use dataset. The data was randomly split into $80\%/10\%/10\%$ training, validation, and test set, respectively. The same structure of the deep convolutional network and training process as the Stockholm and Boden maps were used. The data and the changes to the reasoner is described in more details in Section~\ref{sec:dataucmerced}. The training went through three iterations between the classifier and the reasoner. The per-class and overall classification accuracy on the test set of the first and final round can be seen in Table~\ref{table:confmattable_ucmerced}. All classes get an improved classification accuracy except \textit{non-vegetation ground}. 

The largest confusion, without using a reasoner, is between \textit{vegetation} and \textit{non-vegetation ground}; \textit{building} and \textit{pavement}; \textit{water} misclassified as \textit{airplanes}; \textit{pavement} misclassified as \textit{non-vegetation ground}, \textit{cars}, or \textit{airplanes}; and finally, \textit{cars} and \textit{ships} misclassified as \textit{airplanes}. With the use of a reasoner, the largest improvement is for the reduction of the confusion for the vehicle classes \textit{airplane}, \textit{car}, and \textit{ship}. The confusion between \textit{pavement} and \textit{building} is slightly decreased for both classes. For \textit{pavement}, the confusion for \textit{non-vegetation ground} and \textit{airplanes} is decreased but for \textit{cars} it is increased. The same can be seen for \textit{airplanes}, which greatly reduces the confusion for \textit{pavement} but also increase the confusion for \textit{cars}. However, these increases in confusion is small compared to the largest improvement of removing the confusion between \textit{cars} misclassified as \textit{airplane}. Finally, there is the confusion between \textit{vegetation} and \textit{non-vegetation ground} where the accuracy for \textit{vegetation} is increased but for \textit{non-vegetation ground} is decreased. The reasoner increases the confusion of \textit{non-vegetation ground} as \textit{vegetation}. The reason for this could be due to the broad inclusion of previous classes into \textit{non-vegetation ground} that was merged from bare soil, sand, and chaparral, which is somewhat semantically similar to grass but not trees that the class \textit{vegetation} consists of. A different definition of moving grass to the class \textit{non-vegetation ground} and simply call it \textit{ground} could reduce the confusion between these two classes. 


\begin{table*}[!h]
\centering
\begin{tabular}{c|c c c c c c c c}
     & \multicolumn{8}{c}{Predicted label} \\
     \% & Vegetation & Non-vegetation ground & Pavement & Building & Water & Airplane & Car & Ship \\ \cline{1-9}		 
     \multirow{8}{*}{\rotatebox{90}{Actual label}} 
& \textbf{81.0 (-4.4)} & 7.7 (+4.4) & 3.0 (+0.0) & 6.0 (+1.3) & 0.6 (-0.2) & 0.0 (+0.6) & 1.9 (-1.7) & 0.0 (+0.0) \\
& 13.6 (-5.3) & \textbf{75.2 (+6.6)} & 5.9 (+0.1) & 4.6 (-1.2) & 0.3 (-0.1) & 0.0 (+0.3) & 0.4 (-0.4) & 0.0 (+0.0) \\ 
& 4.1 (-1.8) & 4.8 (+3.4) & \textbf{79.0 (-0.5)} & 5.1 (+0.9) & 0.3 (-0.0) & 0.3 (+3.2) & 4.7 (-4.1) & 1.7 (-1.0) \\
& 5.9 (-2.2) & 1.7 (+2.9) & 5.8 (+2.3) & \textbf{84.5 (-3.2)} & 0.8 (0.5) & 0.1 (+0.9) & 1.2 (-1.1) & 0.0 (+0.0) \\ 
& 5.6 (-0.6) & 3.8 (+1.8) & 0.9 (-0.6) & 0.2 (+0.1) & \textbf{87.1 (-6.4)} & 0.0 (+7.2) & 0.0 (+0.0) & 2.4 (-1.5) \\ 
& 0.0 (+0.0) & 0.4 (+0.1) & 9.5 (+15.8) & 4.8 (-1.5) & 0.0 (+0.2) & \textbf{76.3 (-8.2)} & 9.0 (-6.7) & 0.0 (+0.3) \\ 
& 1.0 (+0.7) & 0.1 (+0.6) & 2.3 (+0.7) & 1.8 (+1.5) & 0.0 (+0.1) & 0.2 (+25.4) & \textbf{94.6 (-31.9)} & 0.0 (+2.9) \\ 
& 0.0 (+0.0) & 0.0 (+0.0) & 0.2 (+0.3) & 0.0 (+0.1) & 1.0 (-0.5) & 0.0 (+7.5) & 0.0 (+0.8) & \textbf{98.9 (-8.2)} \\ 
   \end{tabular}
   \caption{Confusion matrix [\%] on the test set for the classifier that was trained on UC Merced Land Use dataset with the use of the reasoner. The numbers in parenthesis show how the result would change compared to a classifier that did not use a reasoner.}
   \label{table:confmattable_ucmerced}
\end{table*}

\begin{figure*}[ht]
 \centering
\includegraphics[width=0.8\textwidth]{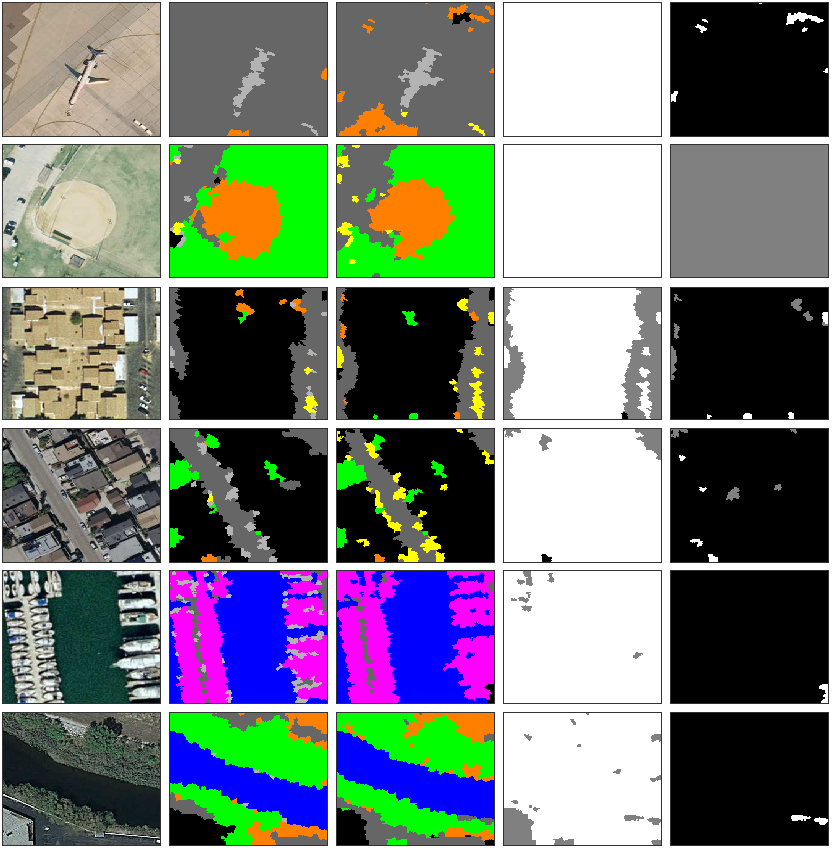}
\caption{Input image (column 1), predictions from classifier without and with using a reasoner (column 2-3, green=vegetation, orange=non-vegetation ground, gray=pavement, black=building, blue=water, yellow=car, purple=ship, light gray=airplane), shadow estimation from reasoner (column 4, gray=undefined, white=not shadow, black=shadow), and height estimation from reasoner (column 5, black=low height, white=large height).}
  \label{fig:ucmerced_results}
\end{figure*}

The input image, predictions, shadow estimation, and height estimation for 6 test images from the UC Merced Land Use dataset can be seen in Figure~\ref{fig:ucmerced_results}. The second column shows the predictions without using a reasoner and the third column shows the predictions when using the reasoner after three iterations of training. The predictions have been averaged within each region to reduce noise. The forth and fifth columns show the shadow and height estimations from the reasoner. From the first four images it can be seen that the reasoner helps to more accurately predict \textit{cars} (a class that should have a small size). It can also be seen that many misclassifications that were predicted as \textit{airplane} have been changed with the use of a reasoner (a class that should have a larger size). There is still some confusion between \textit{pavement} and \textit{non-vegetation ground} that seems to not have been captured semantically in the reasoner. From the last image we see that the reasoner removes some of the confusion between gray roofs from a \textit{building} and \textit{pavement}. 

A summary of the overall classification accuracy for all three data sets with and without the use of a reasoner can be seen in Table~\ref{table:overallacc_stockboden}. The overall classification accuracy is increased for all three data sets when additional information from the reasoner is used. 
\begin{table}[!h]
\centering
\begin{tabular}{c|p{2.4cm}|p{2.4cm}} 
Data set & Overall accuracy [\%] with reasoner & Overall accuracy [\%] without reasoner \\ \cline{1-3}
Stockholm & 86.7 & 81.4 \\
Boden     & 81.3 & 73.2 \\
UC Merced & 83.3 & 77.5 \\
   \end{tabular}
   \caption{Overall classification accuracy for the both classifier trained on only Stockholm or Boden training data for the test data from both cities with and without the use of a reasoner.}
   \label{table:overallacc_stockboden}
\end{table}

\subsection{Evaluation of elevation estimation from the reasoner}
\label{sec:evalreasonelevation}

For data sets that already contain some of the information that the reasoner can provide, it is justified to add it directly as input to the classifier instead of estimating it with the reasoner. One such feature is height information that could come from a Digital Surface Model (DSM), which was available for the maps of Stockholm and Boden, but not for the UC Merced Land Use data set. Table~\ref{table:stockholm_height} shows a comparison of the classification accuracy between a classifier that was trained on RGB and elevation information from a DSM and a classifier that was trained on RGB and elevation information from the reasoner. It can be seen that the classifier that was trained on the ground truth DSM gave slightly better overall classification accuracy. However, the reasoner-provided elevation gives comparable results and is even better for some classes (\textit{vegetation} and \textit{building}). This shows that using a reasoner is a viable replacement for data sets that do not contain elevation information. Furthermore, a reasoner can provide other features, such as shadow. 

\begin{table}[!h]
\centering
\begin{tabular}{l|p{1.9cm}|p{1.7cm}} 
Class                 & Accuracy [\%] with reasoner elevation & Accuracy [\%] with DSM elevation \\ \cline{1-3}
Vegetation            & 74.16 & 69.78 \\
Road                  & 68.37 & 75.76 \\
Building              & 93.33 & 89.80 \\
Water                 & 91.43 & 95.11 \\
Railroad              & 42.12 & 54.88 \\ \cline{1-3}
Mean accuracy         & 80.64 & 82.79
   \end{tabular}
   \caption{Per-class and mean classification accuracy on the testing set on Stockholm when trained on Stockholm RGB and elevation from reasoner or with ground truth DSM.}
   \label{table:stockholm_height}
\end{table}

\subsection{Reasoning for improved learning}
\label{sec:correction}
So far, the reasoner does not correct misclassifications directly but rather influence the learning process of the classifier. The reason for this approach is twofold. Firstly, our objective is to ultimately learn about spatial features about regions, and the reasoner is used to automate the role of a supervisor. Secondly, the reasoner as a referee is also inherently uncertain, and thus may provide several candidate labels for a region. Therefore, the integration between the reasoning and learning architecture has been done in manner described in this paper. 

\section{Discussion and Future Work}
\label{sec:discussion}

This paper has proposed a combination of a deep neural network with an ontological reasoning approach that improves the overall classification accuracy for a semantic segmentation task of RGB satellite images. By applying geometrical processing about spatial features and ontological reasoning based on the knowledge about cities, the semantic referee is able to semantically provide augmented additional input to the classifier which reduces the amount of misclassifications.

It is worth mentioning that this work relies on the suggestions from the semantic referee, which highly depends on the content of the available ontologies. The richer the ontological knowledge, in terms of spatial constraints, the more meaningful the explanations can be expected from the reasoner. It is also important to clarify that we do not categorize this work as a neural-symbolic \textit{integrated} system, since the neural network algorithm is independent of the symbolic reasoning module. However, our proposed architecture can be viewed as a strength since it allows for different types of classifiers to be coupled onto the reasoning system in a straightforward manner, which makes our suggested approach a generic framework.

A future direction is to look into how the the semantic referee could be integrated into the neural network in such a way that the interaction between the two systems is not limited to only the first and last layers but instead is part of the learning process of the hidden layers of the classifier as well. Another interesting future direction is to explore the reverse process, namely how the classifier can enhance the capabilities of the reasoner. 

The source code for this work can be found at: \href{https://github.com/marycore/SemanticRobot}{https://github.com/marycore/SemanticRobot}

\begin{acks}
This work has been supported by the Swedish Knowledge Foundation under the research profile on Semantic Robots, contract number 20140033. The work is also supported by Swedish Research Council.  
\end{acks}

\bibliographystyle{ios1}
\bibliography{iosart2x}

\end{document}